\documentclass[letterpaper, 10 pt, conference]{ieeeconf}      % Use this line for a4 paper

\usepackage{graphics} % for pdf, bitmapped graphics files
\usepackage{epsfig} % for postscript graphics files
\usepackage{epstopdf}
\usepackage{mathptmx} % assumes new font selection scheme installed
\usepackage{times} % assumes new font selection scheme installed
\usepackage{amsmath} % assumes amsmath package installed
\usepackage{amssymb}  % assumes amsmath package installed
\usepackage{url}
\usepackage[font=scriptsize,labelfont=bf]{caption}
\usepackage{color}
\usepackage[backend=bibtex,style=numeric,natbib=true]{biblatex}
\usepackage{mathtools}
\usepackage{pgfplots}
\usepackage{booktabs,color}
\usepackage{subfigure}
\usepackage{float}

\usepackage{subfigure}
\usepackage{float}
\usepackage[utf8]{inputenc}

\pgfplotsset{ % Here we specify options for all figures in the document
  compat=newest, % Which version of pgfplots do we want to use?
  legend style ={font=\footnotesize \sffamily},
  label style = {font=\footnotesize\sffamily},
every tick label/.append style={font=\footnotesize}}
\usepackage{tikz}

\newcommand\inputpgf[2]{{
\let\pgfimageWithoutPath\pgfimage
\renewcommand{\pgfimage}[2][]{\pgfimageWithoutPath[##1]{#1/##2}}
\input{#1/#2}
}}

\newcommand{\e}{{\mathrm e}}

\newcommand{\beq}{\begin{equation}}
\newcommand{\eeq}{\end{equation}}
\newcommand{\bea}{\begin{eqnarray}}
\newcommand{\eea}{\end{eqnarray}}
\newcommand{\beas}{\begin{eqnarray*}}
\newcommand{\eeas}{\end{eqnarray*}}
\newcommand{\ba}{\begin{array}}
\newcommand{\ea}{\end{array}}
\newcommand{\bit}{\begin{itemize}}
\newcommand{\eit}{\end{itemize}}
\newcommand{\ben}{\begin{enumerate}}
\newcommand{\een}{\end{enumerate}}

\title{A Classifiers Voting Model for Exit Prediction\\  of Privately Held Companies}

\author{Giuseppe Carlo Calafiore,  Marisa Hillary Morales,  Vittorio Tiozzo, Serge Marquie% <-this % stops a space
\thanks{G.C. Calafiore is affiliated with the Department of Electronics and Telecommunications, Politecnico di Torino, Torino, Italy, and with IEIIT-CNR, Italy.
M.H. Morales and V. Tiozzo are Master students at Politecnico di Torino. S. Marquie is affiliate with Eurostep Digital. Corresponding author: {\tt\footnotesize giuseppe.calafiore@polito.it}.}%
}

\IEEEoverridecommandlockouts

\begin{document}

\maketitle
\thispagestyle{empty}
\pagestyle{empty}

%%%%%%%%%%%%%%%%%%%%%%%%%%%%%%%%%%%%%%%%%%%%%%%%%%%%%%%%%%%%%%%%%%%%%%%%%%%%%%%%
\begin{abstract}
Predicting the exit (e.g. bankrupt, acquisition, etc.) of privately held companies is a current and relevant problem for investment firms. The difficulty of the problem stems from the lack of reliable, quantitative and publicly available data. In this paper, we contribute to  this  endeavour  by  constructing an exit predictor model based on qualitative data,  which  blends the outcomes of three classifiers, namely, a Logistic Regression model, a Random Forest model, and a Support Vector Machine  model.  The  output  of  the  combined  model is selected on the basis of the majority of the output classes      of the component models. The models are trained using data extracted from the Thomson Reuters Eikon repository of 54697 US and European companies over the 1996-2011 time span. Experiments have been conducted for predicting whether the company eventually either gets acquired or goes public (IPO), against the complementary event that it remains private or goes bankrupt, in the considered time window. Our model achieves  a 63\% predictive accuracy, which is quite a valuable figure for Private Equity investors, who typically expect very high returns from successful investments.
\end{abstract}

%%%%%TO DO
% 1 Add plot

%%%%%%%%%%%%%%%%%%%%%%%%%%%%%%%%%%%%%%%%%%%%%%%%%%%%%%%%%%%%%%%%%%%%%%%%%%%%%%%%
\section{Introduction}
\label{sec:Introduction}

As record amounts of funds are allocated to private equity (PE) investments -- a record breaking \$671 billion in 2017 -- the main issue private equity investors are struggling with has not changed in decades: the absence of transparent, easily accessible valuation-related information. PE investors, looking at possible investments in privately held companies are lacking quantitative information allowing to build their investment case, and are resorting to methods such as portfolio diversification to compensate this lack of company specific information. PE investors critically need structured approaches allowing to infer basic future performance measures such as, the prospect for a company to IPO in the future or the value of a private company as a potential acquisition target. The early capital raising experience of a private company provides some of the first and richest available informational. Typically early stage companies go through successive investment rounds (seed round, series A, series B, etc.) to which various types of investors can participate: Angel investors, Venture Capitalists, Private Equity funds, each with their own focus, expertise and history. Early investors will typically play a significant role in the future development of each company, often taking significant participation, receiving one or several board seats. It is therefore reasonable to believe
 that parameters such as the nature and the composition of a private company early investors contains significant information regarding the future of that company. 

In this paper, three models are developed and used to forecast the future performance of a private company, using as prognostic factors qualitative information available on the company’s first three investment rounds, such as the time lag between the creation of the company and each of the invest- ment rounds. Performance measures are high level measures, such as the future decision to IPO or not, the potential for      a future acquisition or the risk of a potential bankruptcy. In addition to this company specific information, the models developed use also an indicator for market sentiment at the time of each investment, namely the VIX index level.

Automatic classification algorithms are plausible candidates for offering a solution to the problem of predicting private company future performance. This family of algorithms can process large quantities of both qualitative and quantitative data for calibration purposes. They have also proved efficient in similar situations, producing accurate forecasts in areas as diverse as bio-statistics \cite{hardle2006statistical} or corporate finance \cite{hua2007predicting}. 

In our framework, due to the sporadic and qualitative nature of the data available at this stage, it is difficult to expect high accuracy from any forecast. 
%
%In our framework, due to the sporadic and qualitative nature of the available data, we %cannot expect high accuracy prom the predictions. However, successful investments in %PE typically have extremely high rates of return:
%it is because of the combination of information scarcity and potential outsized %returns that private equity investors attribute significant value to performance %forecasting indicators, even with small forecasting power. 
However, contrary to many conventional market strategies, successful investments in PE  typically have extremely high rates of return: 
 Peter Thiel’s investment in Facebook in 2004 (\$500,000) appreciated 693,3\% by the time of Facebook IPO, \cite{facebook_worth}. Softbank investment in Alibaba in 2009 appreciated 290,0\% by the time of Alibaba IPO, \cite{alibaba_worth}. 
% With such returns, forecasting indicators, even with marginal forecasting power, do %provide private equity investors with highly differentiating and value creating  %tools. 
As a result, the combination of information scarcity and potential outsized returns, makes any performance forecasting indicators, even ones with marginal forecasting power, extremely valuable for private equity investors. 

\vspace{.5cm}

In this work we make an initial attempt 
at PE exit prediction, by focusing  on the IPO/Acquired vs Private/Bankrupt classification, which we shall from now on  denote for brevity as IPO and NonIPO, respectively. 
The model we propose is a composite decision model based on three components. Namely,
\begin{enumerate}
    \item Logistic regression (LR), which models the log odds of the IPO and NonIPO
    outcomes via a linear function of the features; \cite{walker1967estimation}.
    \item Random Forest (RF), which is a tree based ensemble method. Using training data, it grows a number of decision trees, each built according to a greedy optimization algorithm. RF then casts the classification label to a new data point according to a voting model across trees (the forest); \cite{breiman2001random}.
    \item Support Vector Machine (SVM), which attempts a geometric separation of the IPO and NonIPO samples via an hyperplane in a high dimensional space.
\end{enumerate}
Since each of these methods has its own strengths and weak points, we observed that better experimental results are obtained by fusing the models into a combined model whose output is given by the majority of the outcomes of the component models, i.e., the IPO or NonIPO prediction is made on the basis of the 2/3 of the component outcomes. 
The algorithms tested in this paper have been already used in market and corporate finance endeavours, see, e.g., \cite{hua2007predicting}, \cite{kumar2006forecasting}. The capabilities  of a Random Forest model have been previously exploited in the context of classification of  private equity data in \cite{bhat2011predicting}.
Moreover, use of SVM and blending of SVM and classification trees has been used for estimation of financial distress in \cite{chen2011predicting}. In general, ensemble methods, from whom fused model descend, have proved to have strong predictive power; \cite{friedman2001elements}.

This paper is organized as follows. In Section~\ref{sec:data} we describe the available data set, in Sections~\ref{sec:iii} and~\ref{sec:predmet} we discuss the
descriptive statistics and prediction metrics used throughout the study. In Section~\ref{sec:components}
we briefly describe the components models and the fused, majority based, model.
Section~\ref{sec:numerics} reports the results of the numerical experiments and a discussion on the performance of the proposed prediction model. Conclusions are finally drawn in Section~\ref{sec:conclusions}.

%\vspace{-2cm}

\section{Input data set}
\label{sec:data}
\begin{figure}[htb]
\centering
\includegraphics[width=0.42\textwidth]{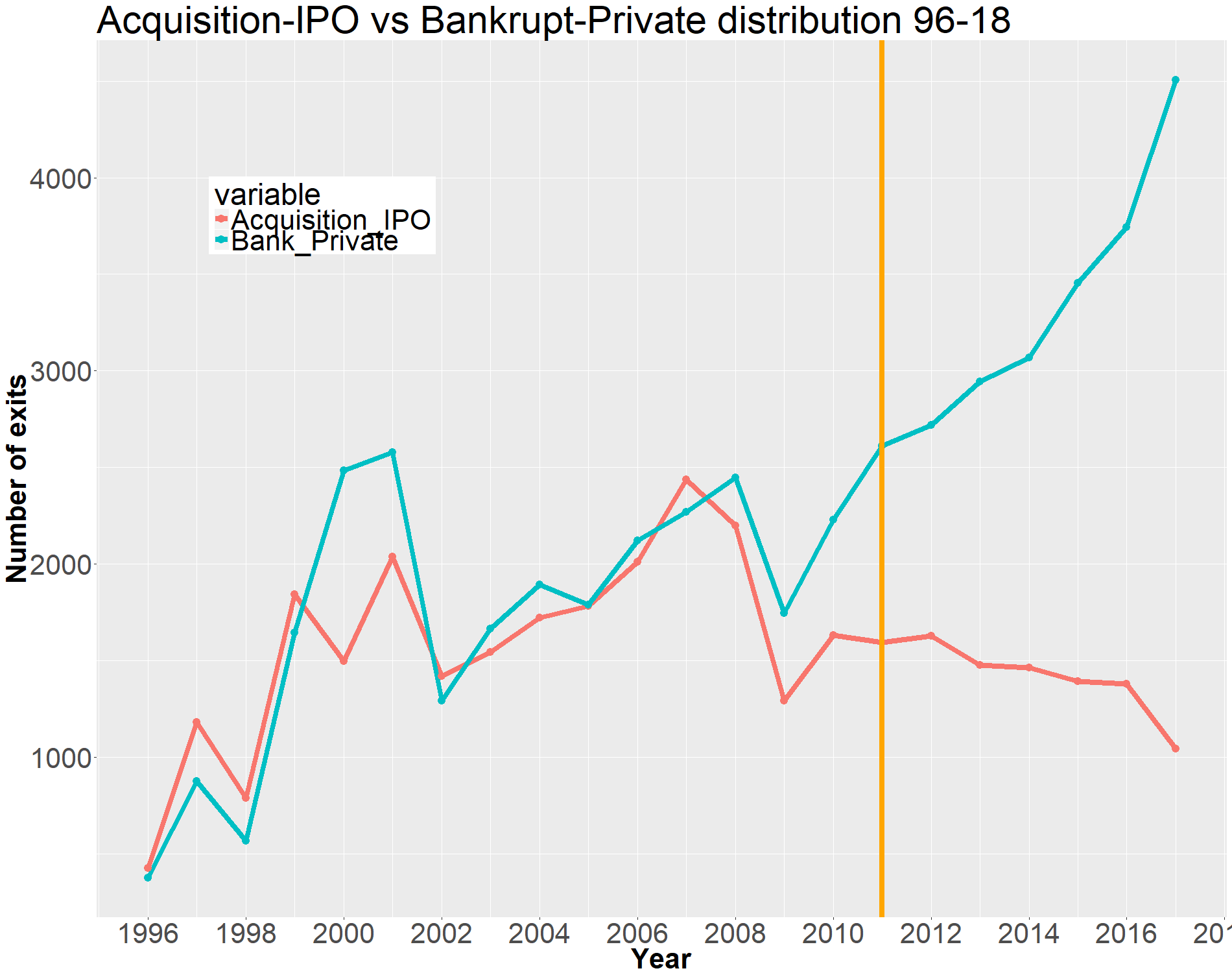}
\caption{\label{fig:distr_bank_year} Exit distribution of Bankrupt/Private and IPO/Acquisition classes in the 1996-2018 period.}
\end{figure}

The data used for this research was extracted from Thomson Reuters Eikon. The full data set contains information on US and European companies between 1996 and 2018. The data set is composed of 83544 companies belonging to 9 different industry sectors. Data is only qualitative: it contains the names of the investors' firms, the date of the investment rounds, the company foundation year, and a public market sentiment indicator, given by the VIX index. 
Only information about the first 3 rounds was retained. The classification output variable (label) was the exit status of the company: IPO, Bankrupt, Merger and Acquisition (M\&A), Leveraged Buyout (LBO), or Private. 
For this paper's purposes, we aggregated the output into two classes, namely IPO (including actual IPOs as well as acquisitions) and NonIPO. The distribution of data for both classes (IPO, NonIPO) across time is shown in Fig.\ref{fig:distr_bank_year}.
%In order to produce a quantitative vector suitable for the predictive algorithms, a transformation of the data found in TE has been applied, and a matrix of companies has been built.
%First we create a ranking of investors based on the frequency and taking into account with whom each investor has invested. Then the matrix is composed by the variables (columns): "Foundation year", "Number of Investors round 1" "Max Investor round 1", "Min Investor round 1", "Mean Investors round 1", "Number of Investors round 2", "Max Investor round 2", "Min Investor round 2", "Mean Investors round 2", "Number of Investors round 3", "Max Investor round 3", "Min Investor round 3", "Mean Investors round 3", "Date round 1", "Date round 2", "Date round 3", "vix1", "vix2", "vix3" for each company (rows).     

\label{sec:iii_1}

\subsection{Descriptive statistics}
A preliminary analysis of the data set reveals a progressive inconsistency in the data in the most recent period (2011-2018). This issue is likely to be caused by a data censoring problem,  i.e., events such as acquisitions and IPO have not yet occurred in the given time frame.
%
%This maybe is due to a data censoring issue. In fact nearly 6\% of the companies were bankrupted %and from 2011 on there were just 22 bankruptcies out of more than 25.000 companies. 
%
%So, even if bankruptcy in the first years of a company is very likely, this class is the minority %one. So, in order to avoid the 2011-2018 data divergence, 
To circumvent this problem, we restricted our analysis 
to the investments in the period between 1996 and 2011, for which the corresponding outcomes are more reliable. {By doing so the number of companies considered decreases from 85344 to 54697.}
The exit distribution for every industry sector is reported in Table~\ref{tab:exit_statistics}.

\begin{table}[htb]
\scalebox{0.89}{
\centering
\begin{tabular}{l|r|r|r|r|r|r}
Sector & Bankrupt & IPO & LBO & M\&A & Private & \textbf{Totals}\\\hline
1 Communications & 481 & 193 & 126 & 1306 & 990 & 3096 \\ 
  2 Computer & 1751 & 644 & 409 & 4687 & 5829 & 13320 \\ 
  3 Electronics & 489 & 325 & 217 & 1152 & 1808 & 3991 \\ 
  4 Biotech/Pharma & 256 & 585 & 294 & 1172 & 1908 & 4215 \\ 
  5 Medical/Health & 328 & 391 & 398 & 1276 & 1846 & 4239 \\ 
  6 Energy & 36 & 183 & 141 & 223 & 624 & 1207 \\ 
  7 Consumer & 506 & 485 & 1629 & 1941 & 3337 & 7898 \\ 
  8 Industrial & 588 & 597 & 1971 & 2450 & 3859 & 9465 \\ 
  9 Other & 661 & 570 & 1011 & 1351 & 3673 & 7266 \\ 
  \textbf{Totals} & 5096 & 3973 & 6196 & 15558 & 23874 &  \textbf{54697} \\  
\end{tabular}}
\caption{\label{tab:exit_statistics}Descriptive statistics of exit distribution, 1996-2011.
}
\end{table}
\label{sec:iii}

\section{Prediction metrics}
\label{sec:predmet}
The predictive performance of the model is quantified by means of five indicators, obtained by the comparison of the algorithm's cast labels and the real ones, on the test data set. 
Considering the ``IPO'' class to be the positive class, we consider the standard indicators
\beas
    \mbox{Prec}\mbox{$_+$} &=& \mbox{Pr}(\mbox{real\ IPO}\ |\ \mbox{classified  IPO})
\label{eq:posprec}
\\
    \mbox{Prec}\mbox{$_-$} &=& \mbox{Pr}(\mbox{real NonIPO}\ |\ \mbox{classified NonIPO})
\label{eq:negprec}\\
   \mbox{Recl}\mbox{$_+$} &=& \mbox{Pr}(\mbox{classified IPO}\ |\ \mbox{real IPO})
\label{eq:recprec}
\\
    \mbox{Recl}\mbox{$_-$} &=& \mbox{Pr}(\mbox{classified NonIPO}\ |\ \mbox{real NonIPO})
\label{eq:recprec}
\eeas
Each indicator is a conditional probability; for example,  the positive precision 
$    \mbox{Prec}_+$ expresses the probability that, given that a company is
classified as IPO by our model, the company will actually be publicly listed or acquired.
For each company the algorithm provides as output the probability of an IPO exit. A threshold $\gamma\in(0,1)$ parameter is then used to decide the class label, i.e., a company is labeled as IPO if its IPO probability is greater than the selected $\gamma$ threshold, and NonIPO otherwise.
The $\gamma$ level is tuned via cross validation,
in order to optimize the predictive performance of the model. 
\label{sec:iv}

\section{Component models and fused model}
\label{sec:components}
The three standard prediction models used in this study are briefly described next.
For all models the training variables $X_1,\ldots,X_n$ 
consist in a transformation of the data described in Section \ref{sec:iii_1}. Namely, for each company we consider (i) a ranking indicator of the importance of the firms that invested in the company, (ii) the dates of the investment rounds, (iii) the company foundation year, 
and (iv) the VIX volatility public index value. Input investment dates consist in the time interval with respect to the company's foundation year.
%Names of the investors' firms is a qualitative variable, so a ranking transformation is %performed. 
%A weighted graph is created: investors are the nodes and the links consist in the number of %investments that two investment firms have made in the same companies. Finally the importance %of an investment firms (i.e. its position within the ranking) is the sum of all its in- %(out-) links.\\
%This could seem a simple ranking method. However, the common sense suggests that if an %investment firm performs a large number of investment, a good number of them has been %successful. Otherwise the firm would have collapsed before long because of bad investments.

\subsection{Logistic Regression}
Logistic Regression (LR) assigns an output probability 
$p(X)$ to a vector of features $X$ according to the model
\begin{equation}
    p(X) = \frac{\e^{\beta_0+ \beta_1X_1 + .. + \beta_nX_n}}{1 + \e^{\beta_0+ \beta_1X_1 + .. + \beta_nX_n} }
\label{eq:logitfunc}
\end{equation}
where $\beta_i$, $=0,\ldots,n$, are the logistic regression coefficient, to be estimated
in the training phase of the model; \cite{walker1967estimation}, \cite{christensen2006log}.

\subsection{Random Forest}
The Random Forest (RF) model is composed by an ensemble (forest) of $N$ classification trees, see, e.g., \cite{breiman2001random}. Each of the $N$ trees is built using training data.
A classification tree takes samples as input and progressively divides them using binary splits, made on a number $m$ of variables and using a greedy algorithm, \cite{venables2002tree}. Each split is based on thresholds on a subset of the sample's features, which are selected randomly. 
Once a number $N$ of trees is grown, the RF predicts a new sample’s class based on the response of the majority of trees. 
%Example: There are 10 trees. If 7 trees say class A and 3 trees class B, %then the object is labeled as class A. \\
RF outputs are class probabilities, one for each observation. 
%According to the classification these are the probabilities for a company %to go IPO or Not.
Setting the threshold $\gamma$ for a label means that at least a proportion $\gamma$
of the trees must agree on that label in order to cast it as the output.
%
%As an example, let's suppose that ACME INC. is injected in the forest. %Than, 30 out of 100 trees vote "\textit{IPO}" relatively to ACME INC, so %its $p_{Forest}$ equals to 30\%. If $\gamma$ is 20\%, than ACME INC. is %forecast as publicly listed-acquired.

\subsection{SVM}
The Support Vector Machine (SVM) is a classifier that attempts geometric separation of the data
points in a high dimensional feature space, see, e.g., \cite{cristianini2000introduction}. SVM builds a surface $Q$ that (softly) separates the samples belonging to the two classes. In order to adapt $Q$ to the data space structure, different \textit{kernels} can be selected, see, e.g.,   \cite{amari1999improving}, \cite{suykens1999least}.
In this paper we used a radial kernel
\begin{equation}
           K(x,y) = \e^{-\alpha\|x-y\|_2^2}, \ \ \alpha > 0.
\end{equation}

\subsection{Fused Model}
\label{sec:v}
A fused model is constructed by 
feeding the same input to the three models described above, and then selecting as output the majority label. This model extends in the idea of the ensemble learning, averaging a response across an ensemble of very different classifiers. 
%In fact, each of the three algorithms is built on very different basis: a likelihood %function, an ensemble of three and a separation surface, respectively. 
In order to describe the voting dynamics among the three component models, we computed the following experimental quantities:
an ``Agree Ratio'' (AR), i.e., the probability (empirical frequency) 
with which 
all three models agree on the same outcome, $\mbox{Pr}(\mbox{Agreement})$; the ``True Agree Ratio IPO'' (TARI), i.e., the probability of correct IPO classification conditional to the three models agreement, $\mbox{Pr}(\mbox{true IPO}|\mbox{Agreement})$; the ``True Agree Ratio NonIPO'' (TARNI) i.e., the  probability of true NonIPO classification conditional to the three models agreement;  
%$P(True \ NonIPO|Agreement)$. 
and a 
probability that one of the methods (LR, RF, SVM) issues the correct classification while being the minority  (TLR.MIN, TRF.MIN, TSVM.MIN, respectively).
Experimental values of these quantities are reported in Section~\ref{sec:fused}.
%``True Agree Ratio RF/LOG/SVM MINority: probability of  RF, LOG, SVM to cast the correct %classification, conditional of being the minority response algorithm $P(M \ is \ Minority \ %Model \ (M_{min}) \ and \ is \ True\ |\ M \ is \ M_{min})$.

\section{Experimental results}
\label{sec:numerics}
Experiments have been conducted using available data in the period 1996-2011.
In order to adjust the slight positive/negative classes unbalancing shown in Figure~\ref{fig:distr_bank_year}, a randomized balancing resampling was implemented. The positive class was sampled from the training set with replacement until the same cardinality of the negative class was reached. Negative class was sampled without replacement.
%A 10 fold cross validation was introduced in every industrial sector analysis.

\begin{figure}
\centering
\includegraphics[width=0.4\textwidth]{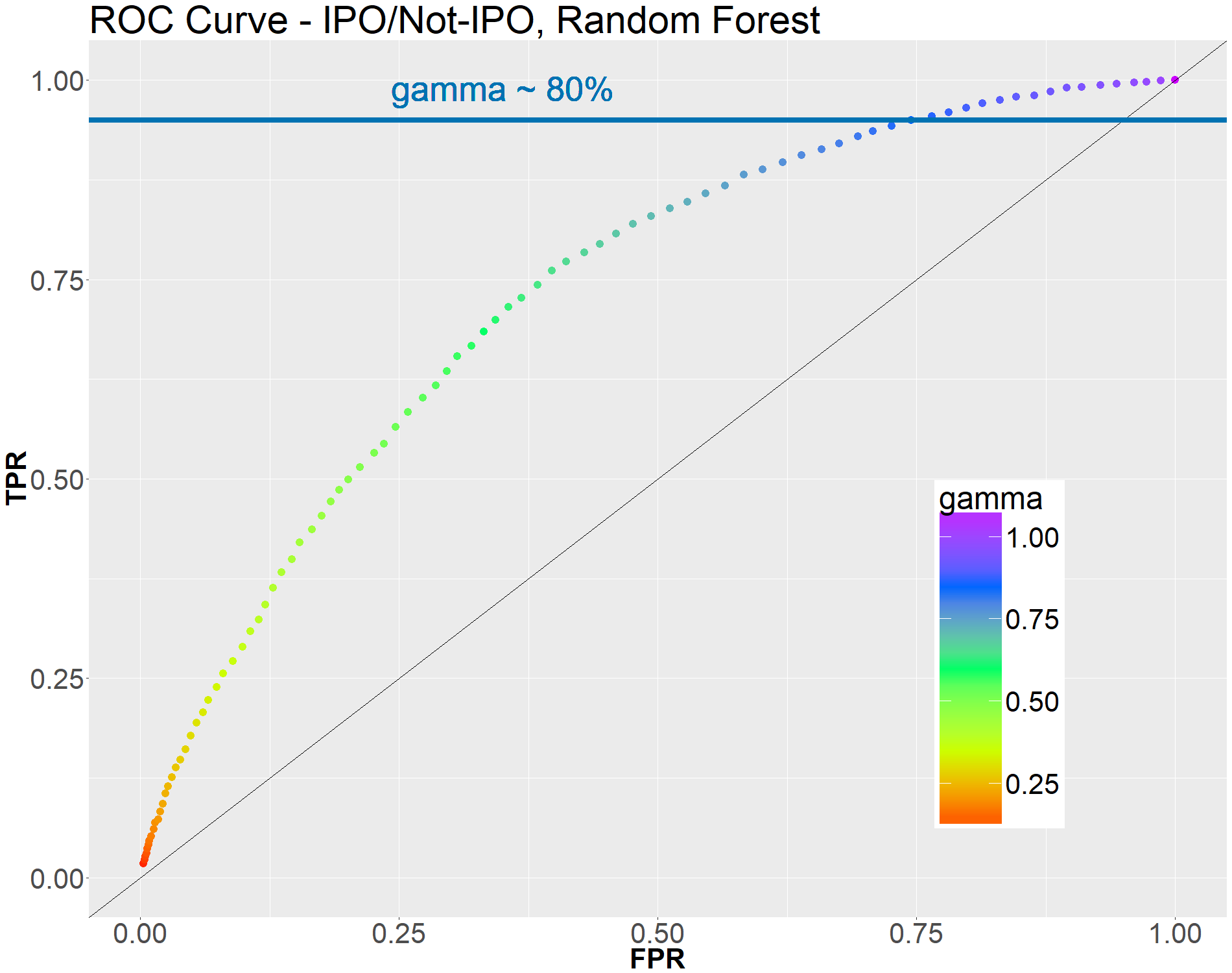}
\caption{\label{fig:roc_rf} Random Forest ROC plot. 1996-2011 investments. Legend represents the probability threshold $\gamma$. Train on 10\% of the data set, randomly sampled. Re-balanced training set. All sectors analysis.}
\end{figure}

Every component algorithm was trained and tested using a 10 fold cross-validation approach. The whole data set was split in 10 sets (or \textit{folds}), randomly sampled and equally sized. Then the algorithm is performed 10 times, using as training set 9 out of the 10 folds. %For each time test set is the fold that is not used as training set and data needed to %compute predictive performance indicators are recorded. 
The predictive performance indicators are then averaged across the 10 measures. 
For each component algorithm, the optimal $\gamma$ value was determined by analyzing the Receiver Operating Characteristic (ROC) plots (Figure~\ref{fig:logistic_ROC}, \ref{fig:roc_rf}, \ref{fig:roc_svm}),  searching for the ``knee'' in the curve that maximizes the true positive rate (TPR) 
 without increasing too much the false positive rate (FPR).
The same new data point will then be presented to the three algorithms, and the three classification labels cast will be compared. The final classification label cast by the fused model will be the one that represents the majority of the three algorithms. 

We used the {\tt R} environment for the whole experimental procedure. Logistic Regression was provided by the default {\tt R}  library, while we used the {\tt randomForest} package for RF, \cite{rfRpackage}, and the {\tt e1071} package for SVM, \cite{svmRpackage}. Training was performed by the {\tt glm}, {\tt randomForest} and {\tt svm} functions respectively. Tuning of SVM was provided by the {\tt tune} function within the {\tt e1071} package. 

\subsection{LR Results}
\begin{figure}[H]
\centering
\includegraphics[width=0.4\textwidth]{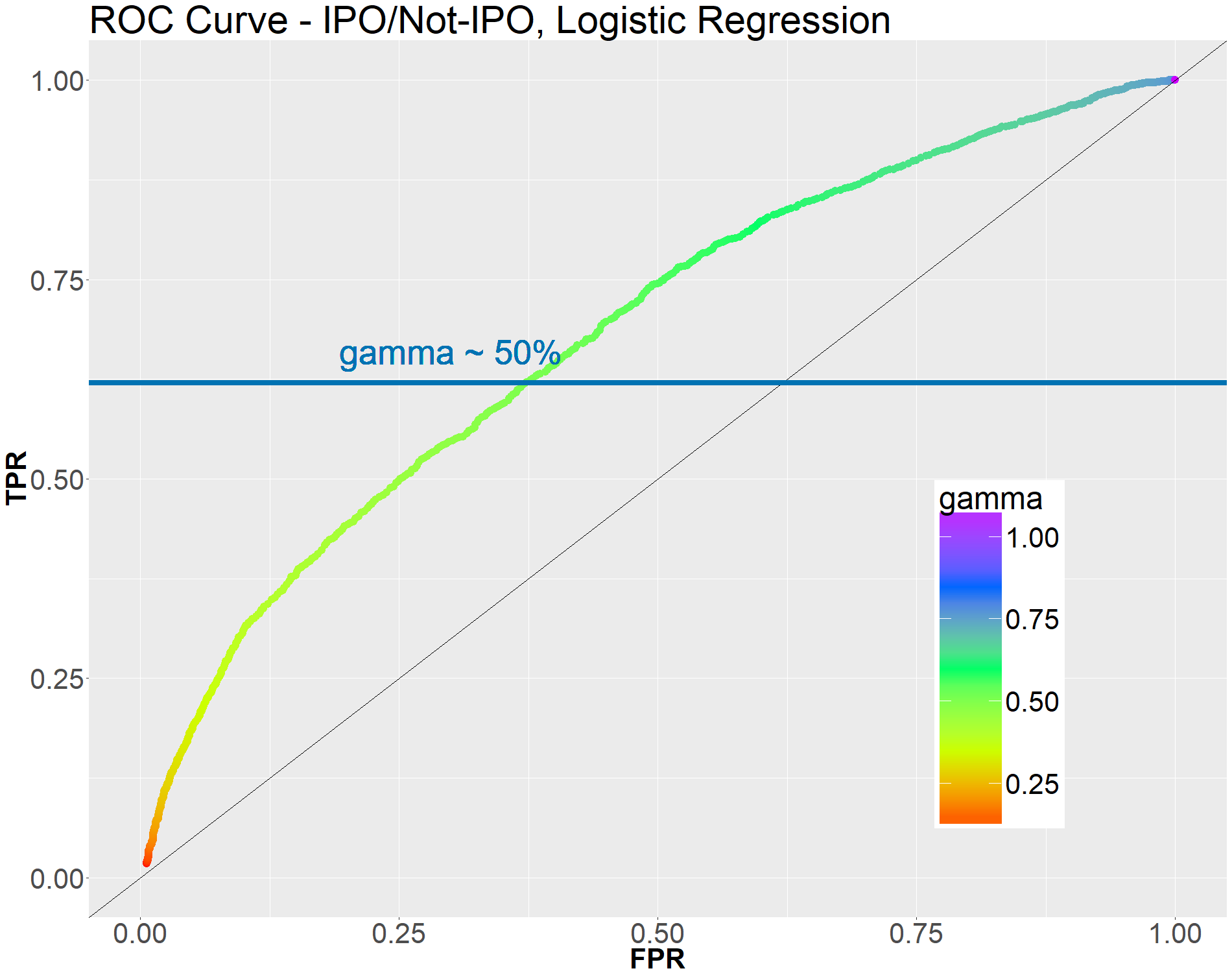}
\caption{\label{fig:logistic_ROC} Logistic Regression ROC plot. 1996-2011 investments. Legend represents the probability threshold $\gamma$. Train on 10\% of the data set, randomly sampled. Re-balanced training set. All sectors analysis.}
\end{figure}

The LR model predictive performance is reported in Table~\ref{tab:res_logit_50} for threshold value $\gamma = 0.5$, and in Table~\ref{tab:res_logit_55} for threshold value $\gamma = 0.55$. The  $\gamma$ value was selected by analyzing the ROC plot in Figure~\ref{fig:logistic_ROC}.
%We have done the same procedure with the SVM and Random Forest ROC %curves.

\begin{table}[H]
\scalebox{0.89}{
\centering
\begin{tabular}{l|r|r|r|r|r}
Sector & Precision+ & Recall+ & Precision- & Recall- & Accuracy \\ 
  \hline
1 Comm. & 0.611 & 0.460 & 0.531 & 0.676 & 0.563 \\ 
  2 Computer & 0.572 & 0.567 & 0.674 & 0.679 & 0.631 \\ 
  3 Electr. & 0.584 & 0.606 & 0.701 & 0.682 & 0.650 \\ 
  4 Bio./Pharma & 0.634 & 0.594 & 0.637 & 0.675 & 0.636 \\ 
  5 Med./Health & 0.606 & 0.523 & 0.599 & 0.677 & 0.602 \\ 
  6 Energy & 0.596 & 0.547 & 0.648 & 0.692 & 0.626 \\ 
  7 Consumer & 0.648 & 0.463 & 0.564 & 0.735 & 0.595 \\ 
  8 Industrial & 0.665 & 0.488 & 0.555 & 0.723 & 0.598 \\ 
  9 Other & 0.596 & 0.465 & 0.685 & 0.787 & 0.657 \\
10 All sectors & 0.615 & 0.510 & 0.622 & 0.717 & 0.620 \\
 \end{tabular}}
\caption{\label{tab:res_logit_50}Logistic Regression predictive performance results. Positive class: IPO. 1996-2011 investments. $\gamma$ = 50\%.}
\end{table}

\begin{table}[H]
\scalebox{0.89}{
\centering
\begin{tabular}{l|r|r|r|r|r}
Sector & Precision+ & Recall+ & Precision- & Recall- & Accuracy \\ 
  \hline
  1 Comm. & 0.590 & 0.669 & 0.571 & 0.487 & 0.582 \\ 
  2 Computer & 0.543 & 0.678 & 0.700 & 0.568 & 0.615 \\ 
  3 Electr. & 0.529 & 0.737 & 0.727 & 0.516 & 0.610 \\ 
  4 Bio./Pharma & 0.586 & 0.713 & 0.658 & 0.522 & 0.615 \\ 
  5 Med./Health & 0.547 & 0.724 & 0.622 & 0.430 & 0.573 \\ 
  6 Energy & 0.575 & 0.654 & 0.676 & 0.598 & 0.624 \\ 
  7 Consumer & 0.603 & 0.682 & 0.611 & 0.527 & 0.607 \\ 
  8 Industrial & 0.612 & 0.703 & 0.597 & 0.497 & 0.606 \\ 
  9 Other & 0.500 & 0.670 & 0.710 & 0.547 & 0.597 \\ 
10 All sectors & 0.563 & 0.698 & 0.659 & 0.518 & 0.603 \\
 \end{tabular}}
\caption{\label{tab:res_logit_55}Logistic Regression predictive performance results.  Positive class: IPO. 1996-2011 investments. $\gamma$ = 55\%.}
\end{table}

\subsection{Random Forest Results}
The performance of the Random Forest model is reported next. Tables \ref{tab:tab:res_rf_50} and \ref{tab:tab:res_rf_80} show the RF predictive performance results for not optimized and optimized $\gamma$, respectively. 

\begin{table}[H]
\scalebox{0.89}{
\centering
\begin{tabular}{l|r|r|r|r|r}
Sector & Precision+ & Recall+ & Precision- & Recall- & Accuracy \\ 
  \hline
 1 Comm. & 0.613 & 0.463 & 0.533 & 0.677 & 0.565 \\ 
  2 Computer & 0.604 & 0.485 & 0.661 & 0.759 & 0.641 \\ 
  3 Electr. & 0.603 & 0.486 & 0.668 & 0.764 & 0.646 \\ 
  4 Bio./Pharma & 0.661 & 0.544 & 0.630 & 0.735 & 0.642 \\ 
  5 Med./Health & 0.635 & 0.453 & 0.592 & 0.753 & 0.607 \\ 
  6 Energy & 0.621 & 0.486 & 0.639 & 0.755 & 0.633 \\ 
  7 Consumer & 0.637 & 0.489 & 0.567 & 0.706 & 0.595 \\ 
  8 Industrial & 0.669 & 0.509 & 0.564 & 0.716 & 0.606 \\ 
  9 Other & 0.604 & 0.498 & 0.696 & 0.779 & 0.666 \\ 
10 All sectors & 0.638 & 0.512 & 0.631 & 0.743 & 0.634 \\
 \end{tabular}}
\caption{\label{tab:tab:res_rf_50}Random Forest predictive performance results. Positive class: IPO. 1996-2011 investments. $\gamma$ = 50\%.}
\end{table}

\begin{table}[H]
\scalebox{0.89}{
\centering
\begin{tabular}{l|r|r|r|r|r}
Sector & Precision+ & Recall+ & Precision- & Recall- & Accuracy \\ 
  \hline
1 Comm. & 0.543 & 0.925 & 0.628 & 0.140 & 0.552 \\ 
  2 Computer & 0.481 & 0.906 & 0.786 & 0.261 & 0.539 \\ 
  3 Electr. & 0.473 & 0.895 & 0.774 & 0.266 & 0.533 \\ 
  4 Bio./Pharma & 0.538 & 0.916 & 0.762 & 0.256 & 0.577 \\ 
  5 Med./Health & 0.510 & 0.931 & 0.697 & 0.150 & 0.531 \\ 
  6 Energy & 0.502 & 0.901 & 0.760 & 0.259 & 0.550 \\ 
  7 Consumer & 0.548 & 0.938 & 0.736 & 0.184 & 0.571 \\ 
  8 Industrial & 0.567 & 0.936 & 0.730 & 0.195 & 0.588 \\ 
  9 Other & 0.489 & 0.924 & 0.870 & 0.346 & 0.579 \\ 
10 All sectors & 0.518 & 0.920 & 0.771 & 0.239 & 0.559 \\ 
 \end{tabular}}
\caption{\label{tab:tab:res_rf_80}Random Forest predictive performance results. Positive class: IPO. 1996-2011 investments. $\gamma$ = 80\%.}
\end{table}

\subsection{SVM Results}
\begin{figure}[h!tb]
\centering
\includegraphics[width=0.4\textwidth]{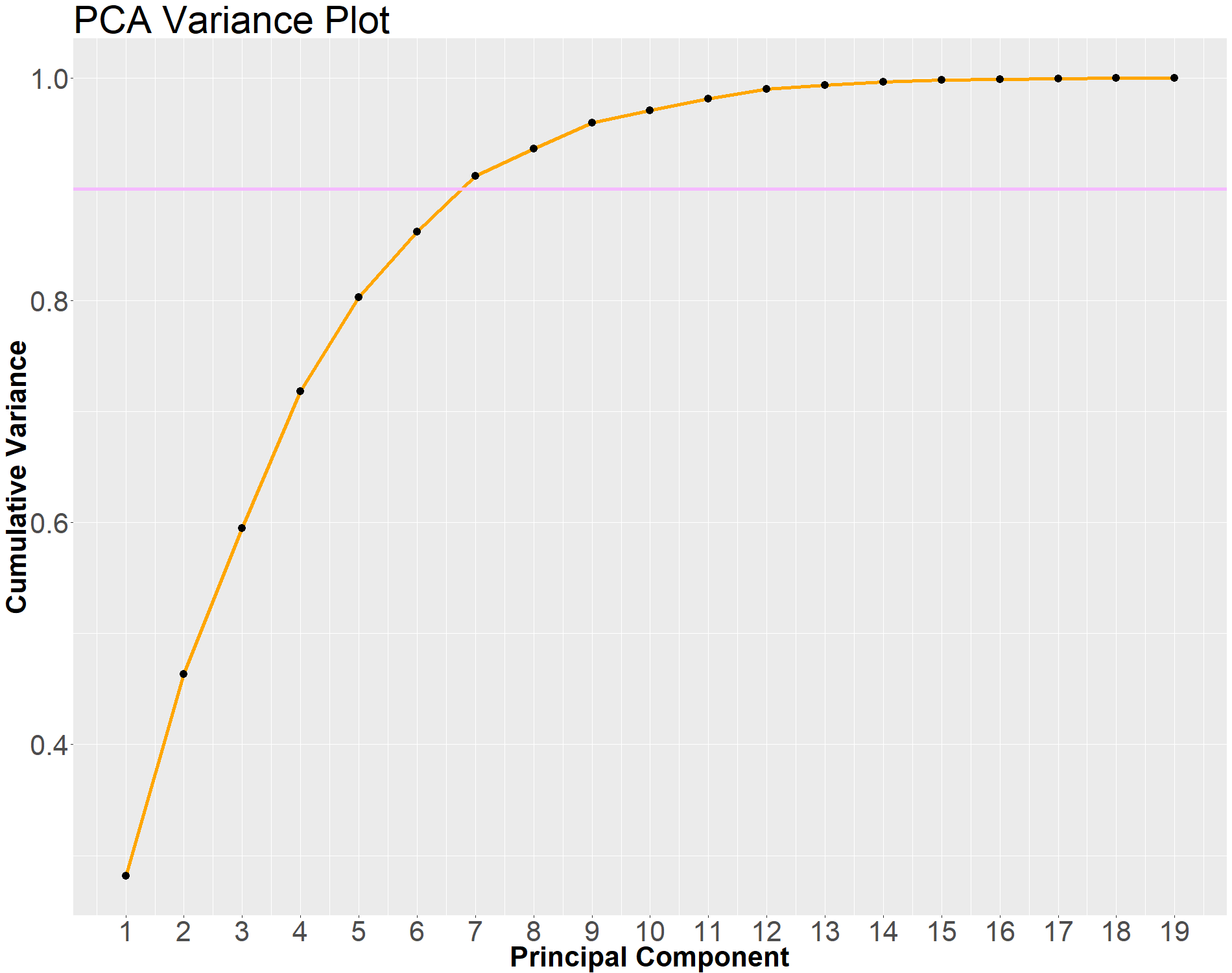}
\caption{\label{fig:pca_plot} PCA cumulative variance plot. At the seventh principal component, the cumulative variance is around 90\% of the total data set.}
\end{figure}

Since the SVM 
turned out to be the most computationally expensive method, we
applied it on a reduced number of features, obtained via Principal Component Analysis (PCA), \cite{wold1987principal}, which was performed  using the {\tt princomp}
 function in {\tt R}. 
Figure~\ref{fig:pca_plot} shows how the seven largest Principal Components describe more than 90\% of variance of the whole original 19-dimensional data space. So, for the SVM tuning and analyses we have selected these first 7 variables.

In the SVM model, we chose a radial kernel with empirically optimized radial parameter $\beta$ and misclassification cost $\nu$ (this is a parameter of the SVM model which accounts for the tradeoff between separation margin and misclassification errors).
Tuning was computationally quite demanding, since it involved multiple SVMs to be trained with different parameters. We alleviated this problem by performing the tuning 
 on a reduced subsample of 1600 data points, along with the PCA dimensional reduction described above. 
We repeated  the tuning session itself 200 times, with a 1600 points sample per session. Then, we computed the median and the most frequent value of tuning parameters found in these 200 tuning sessions, see Table~\ref{tab:svm_emp}). We finally selected $\beta = 0.125$ and $\nu = 0.500$. The algorithm was eventually run on the entire data set, using the optimal parameters determined in the tuning phase.

\begin{table}[H]
\centering
\begin{tabular}{l|r|r}
Statistic & $\nu$ cost & kernel $\alpha$ \\ 
  \hline
Most Frequent & 0.500  & 0.125 \\ 
  Median & 0.516 & 0.129 \\ 
   \end{tabular}
\caption{\label{tab:svm_emp} SVM tuning parameters. }
\end{table}

\begin{figure}[H]
\centering
\includegraphics[width=0.4\textwidth]{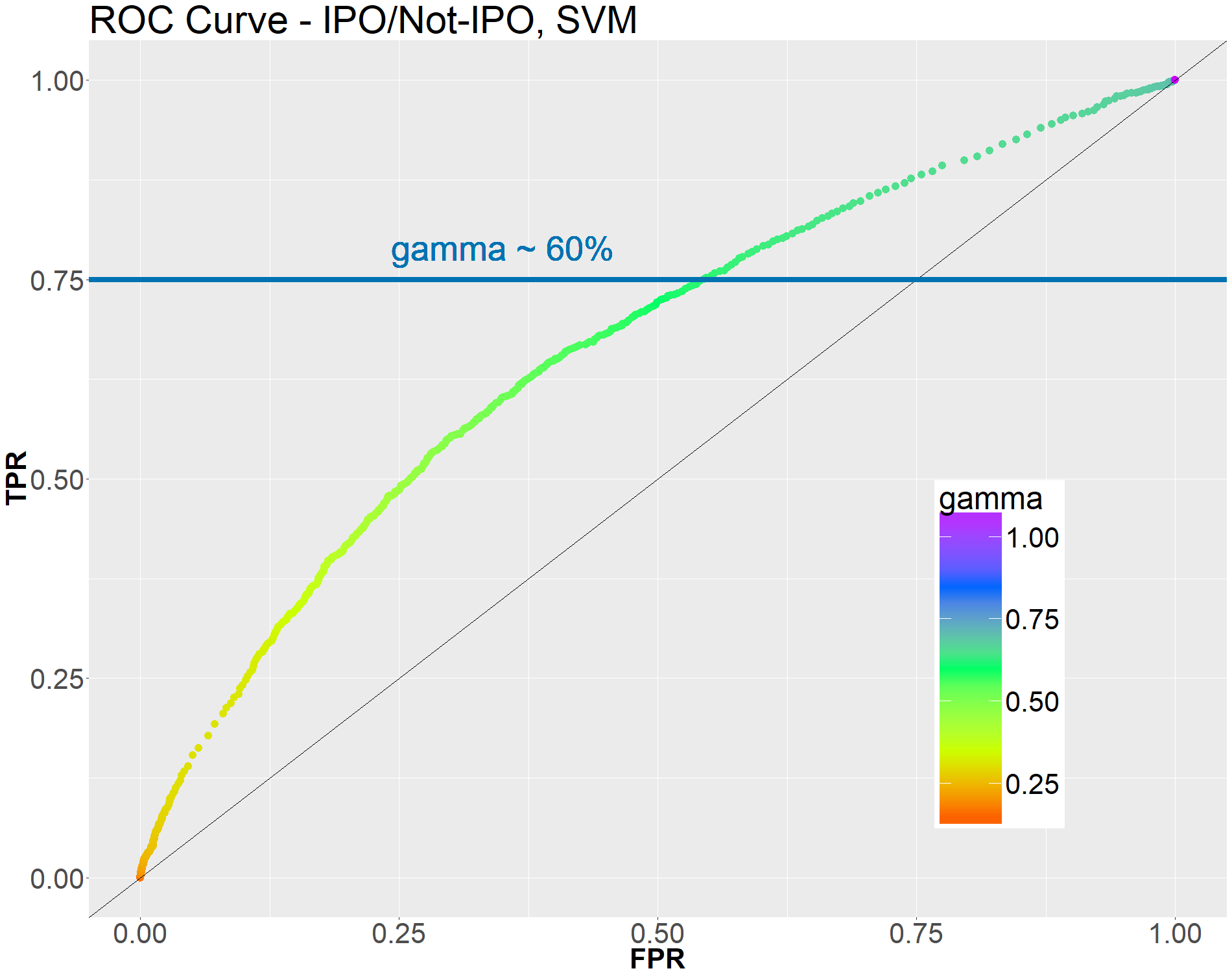}
\caption{\label{fig:roc_svm} SVM ROC plot. 1996-2011 investments. Legend represents the probability threshold $\gamma$. Train on 10\% of the data set, randomly sampled. Re-balanced training set. All sectors analysis. PCA dimension reduction on 7 principal components.}
\end{figure}

The results of the SVM model are reported in Table \ref{tab:svm_50} for $\gamma = 0.5$ and in Table \ref{tab:svm_50_opt} for the optimized $\gamma$'s value found for each sector. Contrarily to the other two algorithms, SVM needs an optimal tuning of every $\gamma$ across different industry sectors. Tuning was performed using histograms of IPO probability, searching for a trade-off between positive recall increase and negative recall decrease.  For the all sectors analysis we relied on the SVM ROC plot in Figure~\ref{fig:roc_svm} for the choice of optimal $\gamma$, which resulted to be approximately equal to $0.6$.

\begin{table} [H]
\scalebox{0.89}{
\centering
\begin{tabular}{l|r|r|r|r|r}
Sector & Precision+ & Recall+ & Precision- & Recall- & Accuracy \\ 
  \hline
1 Comm. & 0.584 & 0.674 & 0.566 & 0.469 & 0.577 \\
2 Computer & 0.557 & 0.651 & 0.698 & 0.609 & 0.627 \\
3 Electr. & 0.558 & 0.674 & 0.716 & 0.605 & 0.634 \\
4 Bio./Pharma & 0.601 & 0.642 & 0.638 & 0.597 & 0.619 \\
5 Med./Health  & 0.572 & 0.627 & 0.610 & 0.555 & 0.590 \\
6 Energy  & 0.548 & 0.664 & 0.664 & 0.549 & 0.601 \\
7 Consumer & 0.600 & 0.617 & 0.585 & 0.568 & 0.593 \\
8 Industrial & 0.631 & 0.629 & 0.585 & 0.587 & 0.609 \\
9 Other & 0.515 & 0.630 & 0.713 & 0.608 & 0.617 \\
10 All sectors & 0.619 & 0.611 & 0.655 & 0.662 & 0.638 \\
 \end{tabular}}
\caption{\label{tab:svm_50}SVM predictive performance results. Positive class: IPO. 1996-2011 investments. Probability Threshold = 50\%.}
\end{table}

\begin{table}[H]
\scalebox{0.80}{
\centering
\begin{tabular}{l|r|r|r|r|r|r}
Sector & Precision+ & Recall+ & Precision- & Recall- & Accuracy & $\gamma$\\ 
  \hline
1 Comm. & 0.579 & 0.898 & 0.666 & 0.238 & 0.594 & 65\%\\
2 Computer & 0.528 & 0.774 & 0.716 & 0.451 & 0.594 & 65\%\\
3 Electr. & 0.561 & 0.834 & 0.771 & 0.461 & 0.630 & 65\%\\
4 Bio./Pharma & 0.518 & 0.920 & 0.714 & 0.190 & 0.545 & 70\%\\ 
5 Med./Health & 0.543 & 0.764 & 0.635 & 0.389 & 0.572 & 60\%\\ 
6 Energy & 0.501 & 0.910 & 0.770 & 0.248 & 0.548 & 65\%\\ 
7 Consumer & 0.547 & 0.901 & 0.669 & 0.211 & 0.565 & 65\%\\ 
8 Industrial & 0.575 & 0.880 & 0.661 & 0.265 & 0.591 & 65\%\\ 
9 Other & 0.419 & 0.931 & 0.733 & 0.127 & 0.452 & 60\%\\ 
10 All sectors & 0.513 & 0.873 & 0.701 & 0.265 & 0.551 &  65\%\\
\end{tabular}}
\caption{\label{tab:svm_50_opt}SVM predictive performance results. Positive class: IPO. 1996-2011 investments. $\gamma$  sector-by-sector optimized.  PCA dimension reduction on 7 principal components.}
\end{table}

\label{sec:vii}

\subsection{Fused Model Results}
\label{sec:fused}
Table~\ref{tab:env1} shows the performance of the fused, majority based, model. 

\begin{table}[H]
\centering
\begin{tabular}{l|r|r|r|r|r}
Sector & Precision+ & Recall+ & Precision- & Recall- & Accuracy\\ 
  \hline
All & 0.627 & 0.512 & 0.634 & 0.734 & 0.631 \\  
  \end{tabular}
\caption{\label{tab:env1}Predictive performance results for the fused, majority-based model. 1996-2011 investments. Positive class: IPO. Probability Threshold = 50\%. PCA dimension reduction on 7 principal components for SVM.}
\end{table}

Table~\ref{tab:env3} reports the  fused model voting dynamics, as described by
the quantities defined in Section~\ref{sec:v}.

\begin{table}[H]

\centering
\begin{tabular}{r|r|r|r|r|r}
AR & TARI & TARNI & TRF.MIN & TLR.MIN & TSVM.MIN \\ 
  \hline
0.691 & 0.671 & 0.671 & 0.508 & 0.420 & 0.477 \\ 
  \end{tabular}
\caption{\label{tab:env3} Fused model parameters.}
\end{table}

The results of fused model shown in Table~\ref{tab:env1} are satisfactory, and in line with the single-models results with $\gamma = 0.5$ (all industrial sectors classification). As stated in Section \ref{sec:Introduction}, for PE investors any
prediction method providing an accuracy that is sensibly better than a fair coin toss is potentially valuable. 

We experimented also by transforming the fused model from a majority to a unanimity model, that is, we issue a IPO label only when {\em all} component model agree on that label. 
In this case, the results are  reported in Table~\ref{tab:env_maj}, and 
improve with respect to the  optimized single models, since Recall- is higher than single models, and Positive+ is strengthened. This is due to the fact that the unanimity  model focuses more on the quality of positive class classification. 

\begin{table}[H]
\centering
\begin{tabular}{l|r|r|r|r|r}
Sector & Precision+ & Recall+ & Precision- & Recall- & Accuracy\\   \hline
All & 0.671 & 0.323 & 0.594 & 0.862 & 0.611 \\ 
  \end{tabular}
\caption{\label{tab:env_maj}Predictive performance results for the fused, unanimity-based model.  1996-2011 investments. Positive class: IPO. Probability Threshold = 50\%. PCA dimension reduction on 7 principal components for SVM. Unanimity dynamic IPO.}
\end{table}

\section{Conclusions}
\label{sec:conclusions}

We presented in this paper an innovative application of machine learning classification models to  forecast the type of exit event for private companies, using some of the rare qualitative data available. 
Performance forecast is indeed the biggest challenge facing private company investors. Contrary to public companies where investors can have access to a plethora of information, information available on private companies is scarce, often inaccurate, and most of the time difficult to access. Therefore, any additional means to acquire a better insight into future performance of private companies is potentially very valuable, particularly considering the very high return on investment provided by early participation in successful ventures.

The analysis showed that standard classifiers (LR, RF, SVM) can provide such an insight, although performance indicators can vary across individual algorithms. A fused model based on these component models
offers the advantage of  balancing the performance of the individual component algorithms, providing more stable and equalized results. 
Using a ``unanimity” version of the fused model provides further improvements, in particular at the level of positive precision and negative recalls, as shown in Table~\ref{tab:env_maj}.
This line of research offers numerous opportunities for further developments, many already underway in collaboration with financial firms active in the private equity market.

\section*{Acknowledgements}
This research was supported by Eurostep Digital.

\small

\printbibliography

{
%\bibliographystyle{./IEEEtran}     % mathematics and physical sciences
%\bibliographystyle{spphys}     % APS-like style for physics
%\bibliography{rndwalk_rebal}}   % name your BibTeX data base

}

\end{document}